\documentclass[letterpaper]{article} % DO NOT CHANGE THIS
\usepackage{aaai2026}  % DO NOT CHANGE THIS
\usepackage{times}  % DO NOT CHANGE THIS
\usepackage{helvet}  % DO NOT CHANGE THIS
\usepackage{courier}  % DO NOT CHANGE THIS
\usepackage[hyphens]{url}  % DO NOT CHANGE THIS
\usepackage{graphicx} % DO NOT CHANGE THIS
\usepackage{natbib}  % DO NOT CHANGE THIS AND DO NOT ADD ANY OPTIONS TO IT
\usepackage{caption} % DO NOT CHANGE THIS AND DO NOT ADD ANY OPTIONS TO IT
\usepackage{algorithm}
\usepackage{algorithmic}

\usepackage{amsfonts}       % blackboard math symbols
\usepackage{nicefrac}       % compact symbols for 1/2, etc.
\usepackage{microtype}      % microtypography
\usepackage{xcolor}         % colors
\usepackage{amsmath}
\usepackage{graphicx}
\usepackage{multirow}
\usepackage{amssymb}
\usepackage{amsthm}
\usepackage{bbding}
\usepackage{pifont}
\usepackage{enumerate}
\usepackage{subfigure}

\usepackage{newfloat}
\usepackage{listings}
\DeclareCaptionStyle{ruled}{labelfont=normalfont,labelsep=colon,strut=off} % DO NOT CHANGE THIS
\floatstyle{ruled}
\newfloat{listing}{tb}{lst}{}
\floatname{listing}{Listing}
\title{Unsupervised Contrastive Learning for Efficient and Robust Spectral Shape Matching}
\author{
    Feifan Luo\textsuperscript{\rm 1},
    Hongyang Chen\textsuperscript{\rm 2}\thanks{Corresponding author.} 
}

\affiliations{
    \textsuperscript{\rm 1}College of Computer Science and Technology, Zhejiang University, Hangzhou 310027, China\\
    \textsuperscript{\rm 2}Research Center for Data Hub and Security, Zhejiang Lab, Hangzhou 311121, China\\
    luoff@zju.edu.cn, dr.h.chen@ieee.org
}

\usepackage{bibentry}
\begin{document}

\maketitle

\begin{abstract}
Estimating correspondences between pairs of non-rigid deformable 3D shapes remains a significant challenge in computer vision and graphics. While deep functional map methods have become the go-to solution for addressing this problem, they primarily focus on optimizing pointwise and functional maps either individually or jointly, rather than directly enhancing feature representations in the embedding space, which often results in inadequate feature quality and suboptimal matching performance. Furthermore, these approaches heavily rely on traditional functional map techniques, such as time-consuming functional map solvers, which incur substantial computational costs. In this work, we introduce, for the first time, a novel unsupervised contrastive learning-based approach for efficient and robust 3D shape matching. We begin by presenting an unsupervised contrastive learning framework that promotes feature learning by maximizing consistency within positive similarity pairs and minimizing it within negative similarity pairs, thereby improving both the consistency and discriminability of the learned features.
We then design a significantly simplified functional map learning architecture that eliminates the need for computationally expensive functional map solvers and multiple auxiliary functional map losses, greatly enhancing computational efficiency.
By integrating these two components into a unified two-branch pipeline, our method achieves state-of-the-art performance in both accuracy and efficiency.
Extensive experiments demonstrate that our approach is not only computationally efficient but also outperforms current state-of-the-art methods across various challenging benchmarks, including near-isometric, non-isometric, and topologically inconsistent scenarios—even surpassing supervised techniques.
\end{abstract}

\begin{links}
    \link{Code}{https://github.com/LuoFeifan77/ContrastiveFMNet}
\end{links}

\section{Introduction}\label{sec:Intro}
Non-rigid deformable shape matching is a fundamental problem in shape analysis and related fields, focusing on establishing meaningful correspondences between shapes. This task has a wide range of applications, including deformation transfer~\cite{sumner2004deformation}, shape interpolation~\cite{eisenberger2021neuromorph}, and statistical shape analysis~\cite{Bogo2014FAUST}.

With recent advancements in deep learning, numerous learning-based approaches have been proposed for non-rigid deformable shape matching, primarily based on the functional map pipeline~\cite{Ovsjanikov2012}. The seminal work FMNet (also known as deep functional maps)\cite{litany2017deep} was the first to leverage learned shape features as optimal descriptors to generate the desired functional maps. Building upon this framework, several deep learning-based methods\cite{eisenberger2021neuromorph, Cao2023, bastian2024hybrid, attaiki2021dpfm} have been introduced to address a variety of shape matching scenarios. However, most existing deep functional maps approaches are still concentrating on optimizing pointwise and functional maps individually or simultaneously, e.g., constraining pointwise maps~\cite{halimi2019unsupervised,Ayguen2020} and functional maps alone~\cite{li2022learning,donati2022deep}, or penalizing both pointwise and functional maps to promote properness~\cite{attaiki2023understanding,Cao2023}. While regularizing functional and pointwise maps is actually to learn informative local feature representations\cite{Cao2023}, these advanced methods overlook the direct supervision of feature representation within the embedding space, resulting in subpar feature representation and inferior matching results. Furthermore, these methods like~\cite{Cao2023,luo2025deep} heavily rely on functional map solvers and multiple auxiliary functional map losses, resulting in complex training paradigms and low computational efficiency. A detailed comparison of methods can be found in Table \ref{tab: others V.S. Ours}.

\begin{table}[h!t]
\centering
\scalebox{1.0}{
\begin{tabular}{lcccccccccccc}
\hline
            & \multicolumn{2}{c}{ProF}   & \multicolumn{2}{c}{Unsup}& \multicolumn{2}{c}{WoS} & \multicolumn{2}{c}{Ol} \\ \hline

FMNet  & \multicolumn{2}{c}{\ding{55}}   & \multicolumn{2}{c}{\ding{55}} & \multicolumn{2}{c}{\ding{55}} & \multicolumn{2}{c}{-} \\ %\hline 
            
GeomFmaps  & \multicolumn{2}{c}{\ding{55}}   & \multicolumn{2}{c}{\ding{55}} & \multicolumn{2}{c}{\ding{55}} & \multicolumn{2}{c}{-} \\ %\hline 

SRFeat   & \multicolumn{2}{c}{\ding{51}}   & \multicolumn{2}{c}{\ding{55}}  & \multicolumn{2}{c}{\ding{51}}  & \multicolumn{2}{c}{-} \\

DUO-FMNet    & \multicolumn{2}{c}{\ding{55}}   & \multicolumn{2}{c}{\ding{51}}  & \multicolumn{2}{c}{\ding{55}}  & \multicolumn{2}{c}{\ding{55}} \\ 

AttentiveFMaps    & \multicolumn{2}{c}{\ding{55}}   & \multicolumn{2}{c}{\ding{51}}  & \multicolumn{2}{c}{\ding{55}}  & \multicolumn{2}{c}{\ding{55}} \\ 

RFMNet  & \multicolumn{2}{c}{\ding{55}}   & \multicolumn{2}{c}{\ding{51}}  & \multicolumn{2}{c}{\ding{51}}  & \multicolumn{2}{c}{\ding{51}} \\ 

ULRSSM  & \multicolumn{2}{c}{\ding{55}}   & \multicolumn{2}{c}{\ding{51}}  & \multicolumn{2}{c}{\ding{55}}  & \multicolumn{2}{c}{\ding{55}} \\ 

DiffZO   & \multicolumn{2}{c}{\ding{55}}   & \multicolumn{2}{c}{\ding{51}}  & \multicolumn{2}{c}{\ding{51}}  & \multicolumn{2}{c}{\ding{55}} \\ 

DeepFAFM  & \multicolumn{2}{c}{\ding{55}}   & \multicolumn{2}{c}{\ding{51}}  & \multicolumn{2}{c}{\ding{55}}  & \multicolumn{2}{c}{\ding{55}} \\ 
Ours & \multicolumn{2}{c}{\ding{51}}   & \multicolumn{2}{c}{\ding{51}}  & \multicolumn{2}{c}{\ding{51}}  & \multicolumn{2}{c}{\ding{51}} \\
\hline
\end{tabular}}
\caption{Method comparison. Our method is the first unsupervised contrastive learning approach specifically for non-rigid deformable 3D shape matching, combining a distinctive set of properties that enhance its performance. Where ProF: \underline{Pro}mote \underline{f}eatures in the embedding space. Unsup: Full \underline{unsup}ervised. WoS: \underline{W}ith\underline{o}ut functional map \underline{s}olver. Ol: \underline{O}nly single unsupervised functional map \underline{l}oss. }
\label{tab: others V.S. Ours}
\end{table}

To overcome the limitations, we introduce the first unsupervised contrastive learning-based method, offering a novel principle for resolving non-rigid 3D shape matching challenges. We first propose the unsupervised contrastive learning framework designed to promote informative feature learning. Positive and negative sets comprising candidate points with high and low similarity, respectively, are defined, derived from ranking vertex embedding similarities. Using these sets, a hybrid similarity generator constructs corresponding similarity pairs, i.e., positive and negative. By minimizing distance between positive pairs while maximizing distance between negative pairs, two novel unsupervised contrastive losses are introduced: cross-contrastive loss, which improves feature representation consistency between shapes, and self-contrastive loss, which enhances feature self-discriminability. 
Then, we build the streamlined functional map learning architecture without the computationally intensive functional maps solver or auxiliary functional map losses, significantly reducing reliance on 
structural assumptions of functional maps (e.g., orthogonality) and minimizing the computational cost.
Consequently, building upon the contrastive learning and functional map framework, our method achieves superior performance in both efficiency and accuracy across a wide range of matching scenarios.
The main contributions are summarized as follows:
\begin{itemize}
    \item The first unsupervised contrastive learning framework for non-rigid deformable 3D shape matching.
    \item A simple yet efficient two-branch architecture without a computationally intensive functional map solver and multiple auxiliary functional losses.
    \item Extensive experiments demonstrate that our approach achieves state-of-the-art performance across challenging benchmarks. 
\end{itemize}
\section{Related Works}\label{sec:RW}
Non-rigid 3D shape matching is a long-standing problem that has been studied extensively over the years. For a comprehensive overview of the field, readers are encouraged to consult surveys such as \cite{sahilliouglu2020recent,deng2022survey}. Here, we review approaches most similar to ours.

\subsection{Axiomatic Functional Map Methods}
A landmark approach to non-rigid deformable shape matching is the functional map framework~\cite{Ovsjanikov2012}, which has inspired numerous subsequent works~\cite{ren2018continuous, Ren2019, Huang2020, Hu2021, ren2021discrete, Gao_2021_CVPR, Gautam2021,fan2022coherent, 2022SmoothNonRigidShapeMatchingviaEffectiveDirichletEnergyOptimization, Donati2022, 2023ElasticBasis}. These methods incorporate functional or map constraints to enhance matching accuracy and robustness. However, axiomatic functional map methods rely heavily on the quality of extrinsic~\cite{salti2014shot} and intrinsic~\cite{sun2009concise,Aubry2011The} handcrafted features. Their performance deteriorates under large-scale deformations, often producing unsatisfactory results. In contrast, our approach directly learns distinguishing features from training data, offering improved accuracy and robustness, especially in challenging matching scenarios.

\subsection{Deep Functional Map Methods}
Unlike axiomatic approaches, deep functional map methods aim to eliminate the reliance on handcrafted features by directly extracting shape features from training data. The pioneering FMNet~\cite{litany2017deep} improved matching results by optimizing the SHOT descriptor~\cite{salti2014shot} using residual MLP layers. Methods like UnsupFMNet~\cite{halimi2019unsupervised} and \textit{Ayguen et al.}\cite{Ayguen2020} enhanced pointwise or functional maps within their loss functions, while other works~\cite{roufosse2019unsupervised,sharma2020weakly} optimized functional maps with properties like bijection, and orthogonality.

Recent two-branch architectures~\cite{Cao2023, Sun_2023_ICCV} supervise both pointwise and functional maps to enhance properness, inspiring subsequent works~\cite{cao2024spectral, cao2024revisiting}. Magnet et al.~\cite{magnet2024memory} proposed a single-branch network to improve functional maps. However, these methods often neglect feature enhancement in the embedding space, limiting feature learning performance. Additionally, they still depend on computationally expensive solvers, and multiple loss functions, resulting in complex networks and reduced efficiency.

In contrast, we propose a novel pipeline that learns informative feature representations for more accurate and robust correspondences while significantly reducing reliance on functional map techniques. Our approach outperforms state-of-the-art methods~\cite{Cao2023, magnet2024memory} across challenging shape-matching scenarios, including those using supervised techniques~\cite{litany2017deep, donati2020deep}, while maintaining high computational efficiency.

\subsection{Contrastive Learning}
Contrastive learning has recently emerged as a dominant technique in self-supervised learning, applied across various domains such as computer vision, natural language processing (NLP), and beyond. The core objective is to bring augmented versions of the same sample closer in the embedding space while pushing apart embeddings of different samples. Follow-up works have extended contrastive learning to structured data, such as graphs~\cite{you2020graph, chen2023attribute, shen2023neighbor}, meshes~\cite{li2022srfeat}, and point clouds~\cite{xie2020pointcontrast, cao2023self, jiang2023non}, among others. In contrast to Li et al.\cite{li2022srfeat}, who employed the PointInfoNCE loss\cite{xie2020pointcontrast}, a common penalty term in contrastive learning that promotes feature learning in a supervised manner and heavily relies on ground truth, we are the first to systematically introduce a comprehensive \textit{unsupervised contrastive learning framework} specifically for non-rigid 3D shape matching. Our approach not only advances the theoretical foundation of shape matching but also outperforms supervised methods, such as~\cite{li2022srfeat}, in terms of matching performance.
\section{Background}
We begin by providing a brief overview of the basic pipeline using deep functional maps and direct interested readers to relevant literature~\cite {ovsjanikov2016computing,donati2020deep,Cao2023} for further details. 

\subsection{Deep Functional Map Pipeline}
Given a pair of non-rigid 3D shapes denoted as ${\mathcal{X}}$ and ${\mathcal{Y}}$ with $|{V_\mathcal{X}}|$ and $|{V_\mathcal{Y}}|$ vertices, respectively. The goal is to compute a high quality dense correspondence between these shapes in an efficient way. The basic learning pipeline estimates a functional map between ${\mathcal{X}}$ and ${\mathcal{Y}}$ using the following five steps.

(1) \textbf{Precompute operators.} Compute the first $k$ eigenfunctions $\Phi_{\mathcal{X}} \in \mathbb{R}^{|{V_\mathcal{X}}| \times k}, 
\Phi_{\mathcal{Y}} \in \mathbb{R}^{|{V_\mathcal{Y}}| \times k} $ eigenvalues $\Lambda_{\mathcal{X}}, \Lambda_{\mathcal{Y}}\in \mathbb{R}^{k \times k} $ in matrix notation via generalized eigendecomposition of the corresponding Laplacian matrices $L_{\mathcal{X}}\in \mathbb{R}^{|{V_\mathcal{X}}| \times |{V_\mathcal{X}}|}, L_{\mathcal{Y}}\in \mathbb{R}^{|{V_\mathcal{Y}}| \times |{V_\mathcal{Y}}|}$, respectively. The Moore-Penrose pseudo-inverse of $ \Phi_{\mathcal{M}}$ is $ \Phi^{\dagger}_{\mathcal{X}}=\Phi^{\mathrm{T}}_{\mathcal{X}} \mathbf A_{\mathcal{X}}$, where $\mathbf A_{\mathcal{M}}$ is the diagonal matrix of lumped area elements, and $\mathrm{T}$ denote transpose operation.

(2) \textbf{Feature extractor.} Compute feature vectors $F_{\mathcal{X}}\in \mathbb{R}^{|V_{\mathcal{X}}| \times d}, F_{\mathcal{Y}}\in \mathbb{R}^{| V_{\mathcal{Y}}| \times d}$ defined on each shape via a feature extractor network $F_\theta$, where $d$ is the dimension of features.

(3) \textbf{Functional map computation.} Compute the functional maps by solving the optimisation problem
\begin{equation}\label{equ:desc and reg}
   \mathbf{C}_{\mathcal{YX}} = \mathop{\arg\min}\limits_{\mathbf{C}\!_{\mathcal{YX}}\!} \left\|   \mathbf{C}_{\mathcal{YX}}\! \Phi_{\mathcal{Y}}\!^{\dagger} \mathbf F_{\mathcal{Y}}\! - \Phi_{\mathcal{X}}\!^{\dagger} \mathbf F_{\mathcal{X}}\! \right\|^{2}_{\mathrm{F}} 
    + \lambda E_{reg}(\mathbf{C}_{\mathcal{YX}}\!), 
\end{equation} 
The first term represents the descriptor preservation term, while the second term, known as the Laplacian commutativity term, enforces structural consistency in the functional map. The parameter 
$\lambda$ acts as a hyperparameter.

(4) \textbf{Functional map penalty.} During training stage, structural regularisation (e.g. orthogonality, bijectivity) is imposed on the functional maps, i.e
\begin{equation}\label{equ: fmap loss}
    L_{fmap} = \theta_{bi} L_{bi} + \theta_{or} L_{or}, 
\end{equation}
where 
\begin{equation}
    L_{bi}  =\left\|  \mathbf{C}_\mathcal{YX} \mathbf{C}_\mathcal{XY} -\mathrm{I}\right\|^{2}_\mathrm{F}, L_{or}  = \left\|  \mathbf{C}_\mathcal{YX}\mathbf{C}_\mathcal{YX}^{\mathrm{T}} -\mathrm{I}\right\|^{2}_\mathrm{F}.  
\end{equation}
To promote proper functional maps, Cao et al.\cite{Cao2023} proposed a coupling loss, i.e. 
\begin{equation}\label{equ: couple loss}
    L_{co}  =\left\| \mathbf{C}_\mathcal{YX} - \Phi_\mathcal{X}^\dagger\Pi_\mathcal{XY}\Phi_\mathcal{Y}\right\|^{2}_\mathrm{F},  
\end{equation}
where the pointwise correspondence matrix $ \Pi_\mathcal{XY}$ can be formulated either as a differentiable doubly-stochastic matrix enabling probabilistic correspondence. 
% Moreover, the bijectivity, orthogonality, and coupling loss terms can be penalized bidirectionally. 

(5) \textbf{Pointwise maps computation.}
At inference time, the pointwise map commonly by nearest neighbor search between the aligned spectral embeddings $\Phi_\mathcal{X}\mathbf{C}_{\mathcal{XY}}^\mathrm{T}$ and $\Phi_\mathcal{Y}$, namely, 
\begin{equation}\label{equ:nnsearch}
    \Pi_{\mathcal{XY}} = NNsearch(\Phi_\mathcal{Y}\mathbf{C}_{\mathcal{YX}}^\mathrm{T},\Phi_\mathcal{X}).
\end{equation}

Most existing deep functional map methods focus solely on optimizing pointwise and functional maps either separately or together, rather than enhancing shape features in embedding space, resulting in less than ideal performance for feature learning and matching results. In contrast, our approach enhances feature representation learning through contrastive learning, resulting in consistent and discriminative features and more accurate correspondence performance. Furthermore, our method eliminates complex and time-consuming components of deep functional maps, such as functional map solvers Eq.~\eqref{equ:desc and reg}, achieving higher computational efficiency than theirs.

\section{Unsupervised Contrastive Learning Framework} 
In this section, we comprehensively introduce an unsupervised contrastive learning framework for enhancing the consistency and discriminative power of feature representations in the embedding space. 

\subsection{Positive and Negative Set}
\newtheorem{PNDef}{Definition}
\begin{PNDef}\label{De:PNDef}
    Considering a vertex $x_i$ from shape $\mathcal{X}$ and vertex sets $\{y_j\}^{|{V_\mathcal{Y}}|}_{j=1}$ from shape $\mathcal{Y}$, 
    calculating the feature similarity scores $ \mathbf{S}_{i,:} =\{\mathrm{sim}(\mathbf{F}_i,\mathbf{F}_j)\}^{|{V_\mathcal{Y}}|}_{j=1}\in\mathbb{R}^{|{V_\mathcal{Y}}|}$, where $\mathbf{F}_i$ and $\mathbf{F}_j$ denote the feature representation of vertex $x_i$ and $y_j$, respectively. Then, selecting the top k largest scores as the positive sets. Formally, this is expressed as 
\begin{equation}
    \mathcal{P}(x_i) = \{ y_j \| y_j \in \mathrm{topk}(\mathbf{S}_{i,:})\} \in \mathbb{R}^{k},
\end{equation}
where $\mathrm{sim}$ denotes a function measuring the similarity between features, and $\mathrm{topk}$ denotes the top-$k$ sampling function. Contrarily, the negative set can be denoted as 
\begin{equation}
    \mathcal{N}(x_i) = \{ y_j \| y_j \notin \mathrm{topk}(\mathbf{S}_{i,:})\} \in \mathbb{R}^{|{V_\mathcal{Y}}|-k}.
\end{equation}
\end{PNDef}
We define the positive set and negative set according to the similarity rank of the feature embeddings. Since the similarity score represents the correspondence probability between shapes, the positive set and negative set actually encode the high and low correspondence probability between the shape vertices, respectively.

\subsection{Hybrid Similarity Generator}
Building on the previously introduced positive and negative sets, we propose a hybrid similarity generator that produces corresponding positive and negative similarity pairs through two main steps: similarity estimation and similarity sampling. The core of the similarity estimation step involves computing the similarity score matrix $\mathbf{S}_{\mathcal{XY}}\in \mathbb{R}^{|V_\mathcal{X}| \times |V_\mathcal{Y}|}$, which measures the similarity between vertex features $\mathbf{F}_\mathcal{X}$ and $\mathbf{F}_\mathcal{Y}$. The similarity metric can be defined using either a distance-based or a cosine-based function. In this work, we adopt the cosine similarity function, defined as follows:
% \begin{equation}\label{equ: compute inter similarity}
%     \mathbf{S}_{\mathcal{XY}} =\cos{(\mathbf{F}_\mathcal{X} / ||\mathbf{F}_\mathcal{X}||, \mathbf{F}_\mathcal{Y}/ ||\mathbf{F}_\mathcal{Y}||)} \in \mathbb{R}^{|V_\mathcal{X}| \times |V_\mathcal{Y}|}
% \end{equation}
% where $||\cdot||$ denotes L2 norms, 
\begin{equation}\label{equ: compute inter similarity}
    \mathbf{S}_{\mathcal{XY}} =\mathbf{F}_\mathcal{X}\mathbf{F}^{\mathrm{T}}_\mathcal{Y} \in \mathbb{R}^{|V_\mathcal{X}| \times |V_\mathcal{Y}|}
\end{equation}
where the features are $L_2$ row-wise normalized before computing dot products. $s_{ij}$ represents the feature similarty between vertex $x_i$ and $y_j$. Clearly, different vertex features result in different similarity score matrices, capturing the relationships between vertex pairs $(x_i,y_j)$ in distinct feature spaces.
% Obviously, different vertex features lead to different score matrices, describing vertex pairs’ relations in different feature spaces.

% After estimating the similarity scores of all vertex pairs, we then conducts a two-stage sampling process involving positive similarity sampling and negative similarity sampling to generate the similarity sequences. Here, we introduce the sampling process based on the similarity matrix $\mathbf{S}_{\mathcal{XY}}$ for a simplified description to generate a positive similarity matrix.

After estimating the similarity scores for all vertex pairs, we perform a two-stage sampling process to construct the positive and negative similarity pairs, respectively.
% we perform a two-stage sampling process—positive similarity sampling and negative similarity sampling—to construct the positive and negative similarity pairs. 
% For clarity, we describe the sampling process based on the similarity matrix $\mathbf{S}_{\mathcal{XY}}$ and use the matrix representation for both positive and negative similarity pairs. 
The elements of the $i$-th row of the similarity matrix $\mathbf{S}_{\mathcal{XY}}$ represent the similarity scores of vertex $x_i$ and all vertices from shape $\mathcal{Y}$. Therefore, computing the positive similarity pairs involves selecting the first 
$k$ largest elements from each row, namely:
\begin{equation}\label{equ: positive similarity}
    \mathbf{S}_{\mathcal{XY}}^{+} = \{s_{ij} |s_{ij} \in \mathrm{topk}(\mathbf{S}_{i,:}), i=1,2,..., |V_\mathcal{X}|\} \in \mathbb{R}^{|V_\mathcal{X}| \times k},
\end{equation}
$\mathbf{S}_{\mathcal{XY}}^{+}$ encodes the set of top $k$ most similar vertex embeddings from shape $\mathcal{Y}$. 

\begin{figure}[h!t]
	\centering
	\includegraphics[width=1.0\linewidth]{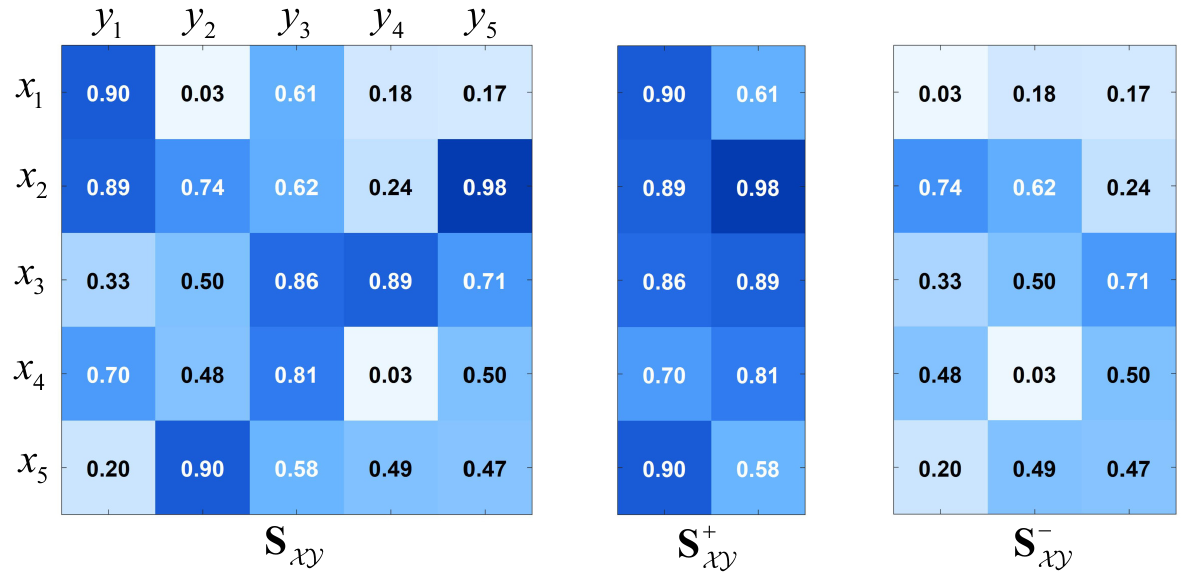}
	\caption{ A visual example illustrating the similarity score $\mathbf{S}_{\mathcal{XY}}$, the positive similarity pair $\mathbf{S}^{+}_{\mathcal{XY}}$, and the negative similarity pair $\mathbf{S}^{-}_{\mathcal{XY}}$. }
	\label{fig2: similarity_vis}
\end{figure}

% As for the negative similarity matrix sampling stage, we have the rest of the similarity matrices $ \mathbf{S}_{\mathcal{XY}}^{R} =\mathbf{S}_{\mathcal{XY}} \setminus \mathbf{S}_{\mathcal{XY}}^{+}\in \mathbb{R}^{|V_\mathcal{X}| \times (|V_\mathcal{Y}|-k)}$, where $\setminus$ symbol represents the set difference operation. Since their similarity scores are below the threshold of top-k selection, we directly regard $ \mathbf{S}_{\mathcal{XY}}^{R}$ as the negative samples. Then, we apply the sampling  function to sample nodes from $\mathbf{S}_{\mathcal{XY}}^{R}$ to construct the negative similarity matrix, i.e., 
For the negative similarity sampling stage, we define the remaining similarity pairs as  $ \mathbf{S}_{\mathcal{XY}}^{R} =\mathbf{S}_{\mathcal{XY}} \setminus \mathbf{S}_{\mathcal{XY}}^{+}\in \mathbb{R}^{|V_\mathcal{X}| \times (|V_\mathcal{Y}|-k)}$, where $\setminus$ symbol represents the set difference operation. 
% Since the similarity scores in $ \mathbf{S}_{\mathcal{XY}}^{R}$ fall below the top-k threshold, we treat them as negative pairs. 
Then we apply a sampling function to select nodes from $ \mathbf{S}_{\mathcal{XY}}^{R}$ to construct the negative similarity pairs, defined as follows:
\begin{equation}\label{equ: negative similarity}
 \mathbf{S}_{\mathcal{XY}}^{-} = \{s_{ij} |s_{ij} \in Sample(\mathbf{S}_{\mathcal{XY}}^{R})\}.
\end{equation} 
Clearly, based on the similarity ranking and sampling function, the hybrid similarity generator effectively divides the feature similarity score $\mathbf{S}_{\mathcal{XY}}$ into the positive similarity pair $\mathbf{S}^{+}_{\mathcal{XY}}$ and the negative similarity pair $\mathbf{S}^{-}_{\mathcal{XY}}$ (see Figure \ref{fig2: similarity_vis}), which capture high-similarity and low-similarity feature pairs, respectively. 

% \begin{equation}\label{equ: compute similarity}
%     \mathbf{S}_{\mathcal{XX}} =\cos{(\mathbf{F}_\mathcal{X}, \mathbf{F}_\mathcal{X})}
% \end{equation}
% and the corresponding positive similarity sampling and negative similarity sampling can be denoted as follows: 
% \begin{equation}\label{equ: positive similarity}
%     v_{\mathcal{X}}^{i,p} = \{v_{\mathcal{Y}}^j |v_{\mathcal{Y}}^j \in Top(\mathbf{S}_i)\}, v_{\mathcal{X}}^{i,n} = V_{\mathcal{Y}} - v_{\mathcal{X}}^{i,p}
% \end{equation}

\subsection{Cross and Self Contrastive Learning Loss}
The core objective of our contrastive loss is to narrow the distance between the positive similarity pairs while widening the distance between the negative similarity pairs in the feature space. Building on the hybrid similarity generator described above, we propose cross-contrastive and self-contrastive loss functions, which respectively enhance the consistency and discrimination of the learned feature representations.

\textbf{Cross-contrastive loss.} The cross-contrastive loss is calculated as follows:
\begin{equation}\label{equ: inter constraive loss}
    L_{cross}(\mathbf{F}_{\mathcal{X}}, \mathbf{F}_{\mathcal{Y}}) = \frac{1}{|V_\mathcal{X}|} \sum_{i} -log \frac{\exp(\hat{s}^{+}_{ij}/\tau_{c})}{\sum^{n_c}_{j=1}{\exp(s^{-}_{ij}/\tau_{c})}},
\end{equation}
where  $s^{+}_{ij} \in \mathbf{S}_{\mathcal{XY}}^{+},  s^{-}_{ij} \in \mathbf{S}_{\mathcal{XY}}^{-}$, $\hat{s}^{+}_{ij} = \frac{1}{p_c}\sum^{p_c}_{j=1}s^{+}_{ij}$, ${p_c}$ means top ${p_c}$ largest scores as the positive pairs, ${n_c}$ denotes the number of negative pairs for each vertex $x_i$, and $\tau_{c}$ denotes the temperature parameter that controls the sensitivity of penalties on positive and negative similarity. Eq.~\eqref{equ: inter constraive loss} enforces the representation of the target features to be close to the central representation of all positive samples and away from all negative samples. Cross-contrastive loss makes the similar features between $\mathbf{F}_\mathcal{X}$ and $\mathbf{F}_\mathcal{Y}$ more similar and the dissimilar features more dissimilar, which encourages feature consistency between shapes, beneficial for downstream shape matching tasks. Moreover, we can build a bidirectional cross-contrastive loss function by exchanging the positions of $\mathbf{F}_{\mathcal{X}}$ and $\mathbf{F}_{\mathcal{Y}}$ in Eq~\eqref{equ: compute inter similarity}.

\textbf{Self-constrastive loss.}
% The previous loss $L_{cross}$ keeps the features $\mathbf{F}_\mathcal{X}$ and $\mathbf{F}_\mathcal{Y}$ consistent, however, it neglects the self-similarity of the shape features, resulting in insufficient distinguishability for features themselves. To this end, we take into account the similarity relationships within the shapes themselves, and introduce the self-contrastive loss to encourage more discriminative features. 
The previously introduced loss $L_{cross}$ enforces similarity consistency between the shape features $\mathbf{F}_\mathcal{X}$ and $\mathbf{F}_\mathcal{Y}$. In addition, to further enhance the discriminability of the features themselves, we incorporate intra-shape similarity relationships and introduce the self-contrastive loss, which encourages the learning of more discriminative and self-aware feature representations.

% encodes the similarity relationship between the vertex features
Consider shape $\mathcal{X}$ as an illustration, the self-similarity score $\mathbf{S}_{\mathcal{XX}}$ can be expressed as
% \begin{equation}\label{equ: compute intra similarity}
%     \mathbf{S}_{\mathcal{XX}} =\cos{(\mathbf{F}_\mathcal{X} / ||\mathbf{F}_\mathcal{X}||, \mathbf{F}_\mathcal{X}/ ||\mathbf{F}_\mathcal{X}||)} \in \mathbb{R}^{|V_\mathcal{X}| \times |V_\mathcal{X}|}.
% \end{equation}
\begin{equation}\label{equ: compute intra similarity}
    \mathbf{S}_{\mathcal{XX}} =\mathbf{F}_\mathcal{X}\mathbf{F}^{\mathrm{T}}_\mathcal{X} \in \mathbb{R}^{|V_\mathcal{X}| \times |V_\mathcal{X}|}.
\end{equation}
For each vertex, we pick the first $p_s$ similar candidates as the positive similarity pairs, and we obtain
\begin{equation}\label{equ: self positive similarity}
    \mathbf{S}_{\mathcal{XX}}^{+} = \{s_{ij} |s_{ij} \in \mathrm{topk}(\mathbf{S}_{i,:}), i=1,2,..., |V_\mathcal{X}|\} \in \mathbb{R}^{|V_\mathcal{X}| \times p_s}.
\end{equation}
Similarly, the negative similarity pairs can be acquired by downsampling the remaining vertex features, formulated as:
\begin{equation}\label{equ: self negative similarity}
 \mathbf{S}_{\mathcal{XX}}^{-} = \{s_{ij} |s_{ij} \in Sample(\mathbf{S}_{\mathcal{XX}}^{R})\},
\end{equation} 
where $ \mathbf{S}_{\mathcal{XX}}^{R} =\mathbf{S}_{\mathcal{XX}} \setminus \mathbf{S}_{\mathcal{XX}}^{+}\in \mathbb{R}^{|V_\mathcal{X}| \times (|V_\mathcal{X}|-p_s)}$. 

Unlike cross-contrastive learning, self-contrastive learning leverages the fact that each vertex's most similar feature is itself. Therefore, there is no need to encourage proximity to itself or to the central representation of the positive pairs. Instead, it is sufficient to push the representation away from the negative pairs. Based on this insight, we omit the attraction to positive pairs and define the self-contrastive loss as follows:
\begin{equation}\label{equ: self constraive loss}
    L_{self}(\mathbf{F}_{\mathcal{X}})  = \frac{1}{|V_\mathcal{X}|} \sum_{i} -log \frac{1}{\sum^{n_s}_{j\neq i}{\exp(s^{-}_{ij}/\tau_{s})}},
\end{equation}
where $s^{-}_{ij} \in \mathbf{S}_{\mathcal{XX}}^{-}$, ${n_s}$ denotes negative sample size, and $\tau_{c}$ denotes the temperature parameter. Similarly, we can also apply the self-contrastive loss $L_{self}$ for shape feature $\mathbf{F}_\mathcal{Y}$.

\begin{figure}[h!t]
	\centering
	\includegraphics[width=0.9\linewidth]{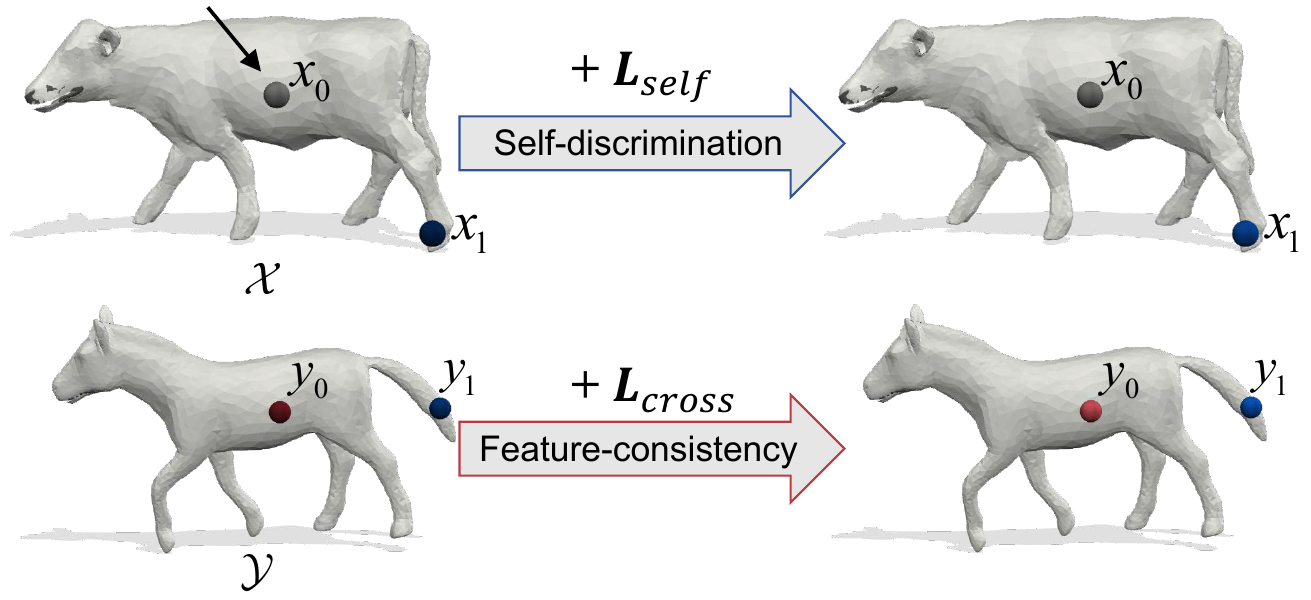}
	\caption{A visual example highlighting the functionality of two contrastive losses. The Euclidean distance between the feature of different points $x_1$ on shape $\mathcal{X}$, and $y_0$, $y_1$ on shape $\mathcal{Y}$, to the source point $x_0$ is computed, where hotter/colder colors mean smaller/larger distances. }
	\label{fig3: loss_vis}
\end{figure}

Interestingly, the self-contrastive loss can be regarded as a special case of the cross-contrastive loss when considering correspondences between identical shapes; however, they serve distinct roles in feature learning. The cross-contrastive loss encourages consistency in feature representations across different shapes, while the self-contrastive loss promotes the self-discriminative ability of vertex representations.  
A visual example is shown in Figure \ref{fig3: loss_vis}. 
Top row: The color of $x_1$ becomes cooler after adding the self-contrastive loss, indicating an increased feature difference between $x_1$ and $x_0$, which demonstrates that the loss promotes self-discrimination by expanding the distance between dissimilar points in the feature space. Bottom row: After adding the cross-contrastive loss, the color of $y_0$ becomes hotter, while the color of $y_1$ becomes cooler, highlighting that the feature difference between $y_0$ and $x_0$ has decreased, and between $y_1$ and $x_0$ has increased. This illustrates that the cross-contrastive loss promotes feature consistency by reducing the distance between similar points and increasing the distance between dissimilar points. 
Together, these two losses complement each other, guiding feature learning from different perspectives and leading to more comprehensive and expressive shape representations. 

\section{Efficient and Robust Unsupervised Contrastive Learning Spectral Shape Matching}
\begin{figure*}[h!t]
	\centering
	\includegraphics[width=0.85\linewidth]{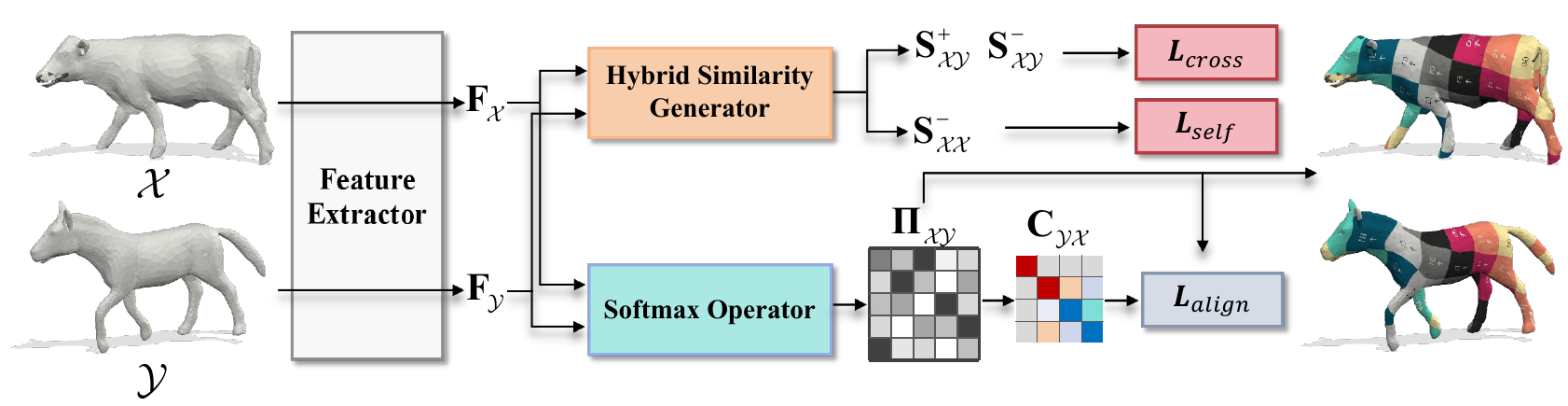}
	\caption{An overview of our method. (1) Feature Extraction: Learned features $\mathbf{F}_{\mathcal{X}}$ and $\mathbf{F}_{\mathcal{Y}}$ are extracted from shapes $\mathcal{X}$ and $\mathcal{Y}$, respectively. (2) Unsupervised Contrastive Learning Branch: The learned features are used to generate positive and negative similarity pairs $\mathbf{S}_{\mathcal{XY}}^{+}$, $\mathbf{S}_{\mathcal{XY}}^{-}$, $\mathbf{S}_{\mathcal{XX}}^{-}$ via the hybrid similarity generator. Two unsupervised contrastive losses Eq.~\eqref{equ: inter constraive loss} and Eq.~\eqref{equ: self constraive loss} are then applied for feature enhancement. 
    (3) Simplified Functional Map Branch: The differentiable pointwise map ${\Pi}_\mathcal{XY}$ is computed using the softmax operator Eq.~\eqref{eq: compute soft map}, and the functional map ${\mathbf{C}}_\mathcal{YX}$ is calculated via spectral basis projection Eq.~\eqref{equ: compute C by Pi}. A functional loss Eq.~\eqref{equ: aglin loss} is constructed to supervise both pointwise and functional map learning.}
	\label{fig4: our_network}
\end{figure*}

In this section, we propose a novel, two-branch unsupervised shape matching approach, which integrates the aforementioned unsupervised contrastive learning framework and the simplified functional map architecture. An overview of the pipeline is illustrated in Figure~\ref{fig4: our_network}.

\subsection{Unsupervised Contrastive Learning Branch} 
The contrastive learning branch follows the previously described unsupervised contrastive learning framework; therefore, we do not reiterate it here but provide guidance on sampling functions and parameter settings. To maintain a simple and efficient shape matching framework, we intentionally omit the sampling functions, although 
incorporating them could potentially improve our method’s performance. Specifically, we set $\mathbf{S}_{\mathcal{XY}}^{-} = \mathbf{S}_{\mathcal{XY}}^{R}, n_c = |V_\mathcal{Y}|-p_c $ for cross-contrastive loss Eq.\eqref{equ: inter constraive loss}, and $\mathbf{S}_{\mathcal{XX}}^{-} = \mathbf{S}_{\mathcal{XX}}^{R}, n_s = |V_\mathcal{X}|-p_s $ for self-contrastive loss Eq.\eqref{equ: self constraive loss}. Additionally, we set $\tau_c = \tau_s$ and $p_c = p_s$ for both losses to further simplify parameter tuning. This approach eliminates the need for designing specialized sampling functions and parameters, demonstrating the robustness of our framework.

\subsection{Simplified Functional Map Learning Branch}
Unlike the functional map learning-based approaches~\cite{li2022learning,Cao2023,luo2025deep} that heavily rely on computationally expensive functional map solver and multiple functional map losses to achieve desirable performances.
Our functional map learning branch is streamlined, comprising only differentiable soft pointwise and functional map computations, along with a single loss term.
For the computation of soft pointwise maps, we forego the higher precision but computationally expensive optimal transport algorithm~\cite{eisenberger2020deep,HU2023101189}. Instead, we utilize the softmax operator to efficiently generate a soft correspondence matrix~\cite{eisenberger2021neuromorph}, namely, 
\begin{equation}\label{eq: compute soft map}
    \Pi_{\mathcal{XY}} = \mathrm{Softmax}(\mathbf{F}_{\mathcal{X}} \mathbf{F}^{\mathrm{T}}_{\mathcal{Y}}/\alpha),
\end{equation}
where the element at position ($i$, $j$) represents the probability of correspondence between the $i$-th point on $\mathcal{X}$ and the $j$-th point on $\mathcal{Y}$, and $\alpha$ is the scaling factor to determine the softness of the correspondence matrix.

Differs from regularized optimization approaches Eq.~\eqref{equ:desc and reg}, an alternative computational strategy establishes an explicit relationship between the functional maps and pointwise correspondence~\cite{Ovsjanikov2012}, namely,  
\begin{equation}\label{equ: compute C by Pi}
\mathbf{C}_\mathcal{YX}=\Phi_\mathcal{X}^\dagger\Pi_\mathcal{XY}\Phi_\mathcal{Y}.
\end{equation}
The functional map $\mathbf{C}_\mathcal{YX}$ is calculated by spectral basis projection Eq.\eqref{equ: compute C by Pi}, rather than the linear system solver Eq.\eqref{equ:desc and reg}, which not only reduces the computational cost but also avoids the instability of the least squares system.

A key component of the deep functional maps framework is the incorporation of loss functions to enforce structural properties of functional maps, such as orthogonality, bijectivity, and so on. Most existing deep functional map methods rely on \textit{multiple} functional map losses to compensate for limitations in their architectures, enabling robust mapping performance across diverse scenarios. 

By effectively promoting feature consistency and discriminability through our contrastive learning framework, conventional bijectivity and orthogonality regularizers can be replaced by the proposed contrastive loss. Consequently, our approach operates without relying on explicit structural constraints, requiring only a single, simple loss function~\cite{HU2023101189} from to promote consistency between the functional map and soft pointwise map, namely:
\begin{equation}\label{equ: aglin loss}
    L_{align}  =  \left\| \Phi_\mathcal{X} - \Pi_\mathcal{XY} \Phi_\mathcal{Y} {\mathbf{C}_\mathcal{YX}}^{\mathrm{T}}\right\|^2_\mathrm{F}.  
\end{equation}

Overall, the total unsupervised loss can be expressed as
\begin{equation}\label{equ: total loss}
    L_{total}  = \theta_{cross} L_{cross} + \theta_{self} L_{self} +  \theta_{align}L_{align},
\end{equation}
where $\theta$ denotes the corresponding weight. 

Finally, we adopt the same strategy as ULRSSM~\cite{Cao2023} to recover the pointwise map during inference, ensuring consistency and accuracy in the final matching results.
\section{Experiments and Results}
\subsection{Baselines}
We extensively compare our method with existing non-rigid deformable shape matching methods, which we categorize as follows:
\begin{itemize}
    \item \textit{Axiomatic approaches}, including ZoomOut \cite{Melzi2019}, Smooth Shells \cite{eisenberger2020smooth}, DiscreteOp \cite{ren2021discrete}, and MWP \cite{Hu2021}.
    \item \textit{Supervised approaches}, FMNet~\cite{litany2017deep}, GeomFmaps~\cite{donati2020deep},SRFeat-S~\cite{li2022srfeat}.
    \item \textit{Unsupervised approaches}, 
    including UnsupFMNet~\cite{halimi2019unsupervised}, SURFMNet~\cite{roufosse2019unsupervised}, Deep Shells~\cite{eisenberger2020deep}, 
    NeuroMorph~\cite{eisenberger2021neuromorph}, 
    DUO-FMNet~\cite{donati2022deep},  AttentiveFMaps~\cite{li2022learning}, RFMNet~\cite{HU2023101189},
    ULRSSM~\cite{Cao2023}, DiffZO~\cite{magnet2024memory}, HybridFMaps~\cite{bastian2024hybrid}, and DeepFAFM~\cite{luo2025deep}. 
    Moreover, ULRSSM introduces a fine-tuning technique called Test-Time Adaptation (TTA), which individually adjusts the network parameters for each test pair during inference. We extend ULRSSM by incorporating TTA, denoted as ULRSSM(+TTA), and perform a comprehensive comparison with this enhanced version. For HybridFMaps~\cite{bastian2024hybrid}—another approach that employs fine-tuning to enhance performance was excluded from comparison. Finally, we will not hightlight supervised methods and fine-tuned matching results because the former relies on ground truth and the latter requires training the network on a test set, and the purpose of using these methods as baselines is to further demonstrate that our methods require neither labeling nor fine-tuning, and still achieve superior performance, even better than them.     
\end{itemize}

%对另一个采用微调提升性能的方法HybridFMaps, 我们没有报告它原论文中的微调结果
% HybridFMaps~\cite{bastian2024hybrid}, the fine-tuned results reported in its original paper are not included in our comparison.

\subsection{Results}
\begin{table*}[h!t]
\centering
\scalebox{0.85}{
\begin{tabular}{lrrrrrrrrrrrrr}
\hline
Train               & \multicolumn{3}{c}{F} & \multicolumn{3}{c}{S} & \multicolumn{2}{c}{F} & \multicolumn{2}{c}{S} & \multicolumn{1}{c}{\multirow{2}{*}{SMAL}} & \multicolumn{2}{c}{DT4D-H} 

\\ \cline{2-11} \cline{13-14}
Test                & F        & S   & S19    & F        & S   & S19    & F\_a      & S\_a  & F\_a    & S\_a  & \multicolumn{1}{c}{}                      & intra-class  & inter-class 

\\ \hline
                    & \multicolumn{13}{c}{Axiomatic Methods}          \\
ZoomOut             & 6.1         & 7.5   & -     & 6.1         & 7.5 & -  & 8.7         & 15.0        & 8.7         & 15.0  & 47.7                                      & 4.0          & 29.0      \\
SmoothShells        & 2.5         & 4.7   & -     & 2.5         & 4.7  & -    & 5.4         & 5.0         & 5.4         & 5.0    & 34.9                                   & 1.1          & 6.3   \\
DiscreteOp          & 5.6         & 13.1   & -     & 5.6        & 13.1 & -  & 6.2         & 14.6       & 6.2       & 14.6   & 36.1                                       & 3.6          & 27.6         \\
MWP                 & 3.1         & 4.1    & -     & 3.1          & 4.1  & -     & 8.2         & 8.7       & 8.2         & 8.7    & 20.9                                   &   1.7           &   25.4        \\ \hline
                    & \multicolumn{13}{c}{Supervised Methods}   \\                    
FMNet               & 11.1         & 30.0    & -     & 33.0          & 17.0  & -   & 42.0         & 43.0     & 43.0          & 41.0    & -                  & 9.6 
                    & 38.0    \\             
GeomFmaps           & 2.6         & 3.4    & 9.9     & 3.0          & 3.0  & 12.2   & 3.2         & 3.8    & 8.4     & 3.1     & 4.3                                   & 2.1  
                    & 4.1      \\
SRFeat-S              & 1.1         & 3.9    & 13.1     & 2.5          & 2.2  & 8.9    & 2.3         & 4.0    & 3.2     & 2.1    & 3.7                                   & 2.4  
                    & 5.0     \\
\hline
                    & \multicolumn{13}{c}{Unsupervised Methods}                                                                     \\
Deep Shells         & 1.7         & 5.4  & 27.4     & 2.7         & 2.5    & 23.4   & 12.0        & 16.0       & 15.0         & 10.0    & 21.4                                    & 3.4          & 31.1  \\
DUO-FMNet           & 2.5         & 4.2   & 6.4     & 2.7        & 2.6  & 8.4    & 3.0         & 4.4        & 3.1         & 2.7     & 4.8                                   & 2.6          & 15.8  \\
AttentiveFMaps      & 1.9         & 2.6  & 6.4      & 2.2        & 2.2     & 9.9  & 2.4         & \underline{2.8}        & \underline{2.5}         & 2.3   & 4.4                                      & 1.7          & 11.6        \\
RFMNet               & 1.7         & \textbf{2.3}        & 6.3         & \textbf{1.7}     & 2.1      & 6.9  & 3.6         & 2.6        & 3.6         & 3.9       & 4.4                    & 1.5          & 13.9    \\
ULRSSM         & {1.6}         & 6.7   & 14.5     &  4.8       & {1.9}   & 18.5   & {2.5}         & 8.9        & 7.0          & {1.9}  &4.5                                      & 0.9          & 5.2       \\
ULRSSM(+TTA)        & 1.6         & 2.2   & 5.7     & 1.6         & 1.9   & 6.7  & 1.9         & 2.4        & 2.1         & 1.9    & 4.2         & 0.9        & 4.1    \\
DiffZO                & 1.9    & \underline{2.4}   & \textbf{4.2}   & 1.9       & 2.4   & \underline{6.9}  & 2.2         & 3.8        & 2.7        & 2.4   & 4.3               & {1.8}                & {4.1}    \\
HybridFMaps & \textbf{1.4}    & 4.2   & 9.5   & 2.3       & \textbf{1.8}   & 13.0 & 2.0        &  4.6       & 3.4      & \textbf{1.8}   & \textbf{3.5}             & 1.0                & \underline{3.9}      \\
DeepFAFM                & 1.6    & 2.7   & 7.0   & 1.9       & \underline{1.9}   & 7.9 & \underline{2.0}         & 2.9        & 2.6        & \underline{1.9}   & {3.9}              & \underline{0.9}                & 4.2      \\
Ours                & \underline{1.6}    & 2.7   & \underline{5.8}   & \underline{1.8}       & 2.1    & \textbf{5.8}  & \textbf{1.9}         & \textbf{2.6}       & \textbf{2.2}        & 2.1  & \underline{3.6}                                       & \textbf{0.9}          & \textbf{3.9}     \\ 
\hline
\end{tabular}}
\caption{ Evaluating the matching results across various benchmarks, including near-isometric shape matching, cross-dataset generalization, anisotropic meshing, and non-isometric shape matching, respectively. The numbers in the table are mean geodesic errors ($\times 100$). \textbf{Bold}: Best. \underline{Underline}: Runner-up.}
\label{tab: near and non-iso}
\end{table*}

Extensive experimental results across multiple datasets, including challenging non-isometric ones, are presented. We use the mean geodesic error~\cite{Kim2011BIM} to evaluate correspondence accuracy, with all results multiplied by 100 for better readability.

\textbf{Near-isometric matching}. 
We evaluate our method on the remeshed versions~\cite{ren2018continuous} of the standard benchmarks FAUST and SCAPE (referred to as F and S, respectively), which are more challenging than the original datasets. FAUST consists of 100 human shapes, representing 10 different people in 10 different poses, and is split into 80 for training and 20 for testing. SCAPE contains 71 human shapes, depicting the same person in various poses, and is divided into 51 for training and 20 for testing. 

The results of these benchmarks are provided in Table \ref{tab: near and non-iso}, where our method is compared with current state-of-the-art axiomatic, supervised, and unsupervised learning approaches. The results indicate that our method performs better than the previous state-of-the-art axiomatic, supervised, and unsupervised methods, such as GeomFmaps and DeepFAFM on FAUST, and achieves comparable results on SCAPE.

\textbf{Cross-dataset generalization.} 
To evaluate the generalization performance of our method, we train networks on remeshed datasets and test them on another remeshed dataset, i.e., training on F and testing on S, and vice versa. Moreover, we use the more challenging SHREC'19 (S19 for short) dataset~\cite{melzi2019shrec} exclusively as a test set. This dataset consists of 44 human shapes, providing a rigorous benchmark for our approach. 

As shown in Table \ref{tab: near and non-iso}, the quantitative results demonstrate our approach outperforms others in most settings. Nevertheless, the existing cutting edge supervised~\cite{li2022srfeat} and unsupervised approach~\cite{Cao2023,bastian2024hybrid} suffers from huge performance drops when testing on the cross-dataset generalisation datasets (e.g., training on F and testing on S19), which substantially demonstrates its inadequate generalization ability compared to existing unsupervised learning-based methods. Additionally, our method achieves superior performance even compared with the fine-tuning matching results of ULRSSM. Clearly, the results also show that the superior performance of ULRSSM heavily relies on the test-time adaptation process, our method is more robust than it.

\textbf{Matching with anisotropic meshing.}
To evaluate robustness across different discretizations, we train networks on remeshed datasets and test them on anisotropic remeshed versions (denoted F\_a and S\_a, respectively), which feature different mesh connectivity compared to the original datasets.

The results presented in Table \ref{tab: near and non-iso} show that our method achieves the best performance with state-of-the-art in most settings and demonstrates greater resilience to changes in triangulation. For instance, when training on S and testing on F\_a, other competitors like HybridFMaps experience significant performance declines, often overfitting to mesh connectivity and producing inaccurate predictions. In contrast, our method maintains strong robustness to varying mesh connectivity and consistently surpasses the current state-of-the-art methods. Furthermore, we achieve performance nearly identical to fine-tuning results, which further underscores the robustness of our approach compared to existing learning-based methods.

\textbf{Non-isometric shape matching.}
In the evaluation of non-isometric shape matching, our approach undergoes rigorous testing across two non-isometric datasets: SMAL~\cite{zuffi20173d} and DT4D-H~\cite{2022SmoothNonRigidShapeMatchingviaEffectiveDirichletEnergyOptimization}, which are significantly more challenging than previous datasets, such as SHREC'19~\cite{melzi2019shrec}, due to the large deformations present in the shapes. The SMAL dataset consists of 49 shapes representing four-legged animals from eight species, divided into a training set of 32 instances and a testing set of 17 instances. In contrast, DT4D-H includes nine classes of humanoid shapes, with a training-testing split of 198 and 95 instances, respectively. 

As shown in Table \ref{tab: near and non-iso}, our method consistently achieves excellent performance across most settings. On more challenging SMAL, our method surpasses existing supervised approaches and achieves performance comparable to HybridFMaps—which integrates both extrinsic and intrinsic information, while other purely state-of-the-art intrinsic spectral techniques like DiffZO and DeepFAFM are unable to attain such results. For the most challenging DT4D-inter dataset, our approach outperforms all baselines, clearly demonstrating the superiority of our method.

For additional comparison results, including topological noise matching, runtime comparisons, and so on, please refer to the supplementary materials.
\section{Conclusion}
We introduce a novel unsupervised contrastive learning-based approach for non-rigid deformable 3D shape matching, comprising the unsupervised contrastive learning framework and the simplified functional map learning module. Extensive experiments demonstrate that our method achieves state-of-the-art performance in both matching results and computational efficiency, establishing a new benchmark for unsupervised 3D shape matching. Additionally, its streamlined design positions it as a foundational framework for future research. 

While effective, our method does not yet support partial shape matching or severely non-isometric deformations. A promising research direction involves integrating our framework with explicit spatial deformation methods~\cite{eisenberger2020deep,eisenberger2021neuromorph,cao2024spectral} to enhance robustness across diverse shape matching scenarios.

\section{Acknowledgments}
We extend our gratitude to Dongliang Cao for his generous assistance. This work is supported in part by National Key Research and Development Program of China 2023YFB4502400, in part by National Natural Science Foundation of China under Grant 62271452. 

% \bigskip
% \noindent Thank you for reading these instructions carefully. We look forward to receiving your electronic files!

\bibliography{aaai2026}

\clearpage

\clearpage
\appendix

\maketitlesupplementary

In this supplementary document, we first provide the implementation details of our method in Section A. Next, we present experiments evaluating the robustness of our method against topological noise in Section B. Section C includes an analysis of the training and inference times across datasets with varying vertex resolutions. In Section D, we perform an ablative study to investigate the impact of the cross-similarity threshold $p_c$ and the self-similarity threshold $p_s$. Section E evaluates the significance of the two contrastive losses in our framework. Eventually, in Section F, we present additional qualitative results of our method.

\section{A. Implementation Details}\label{sec: Imp}
We use DiffusionNet~\cite{sharp2022diffusionnet} as the feature extractor with its default settings, which employs 16-dimensional HKS features~\cite{sun2009concise} as input and generates 128-dimensional learned features for the network. For pointwise map computation, we set $\alpha=0.07$. For functional map computation, we use truncated eigensystems with $k=200$. Regarding our unsupervised loss, we empirically set $\theta_{cross} = 1$, $\theta_{self} = 0.1$, and $\theta_{align} = 1$, respectively. Additionally, we set $\tau_c = \tau_s=1.0$, and configure the number of cross and self-similarities as follows: 
$p_c = p_s =30$ for near-isometric matching, anisotropic meshing, and generalization, $p_c = p_s =10$ for non-isometric matching, and topological matching settings. For training, we use the Adam optimizer~\cite{Kingma15} with a learning rate of $0.001$ for all learning parameters. 

\begin{table}[h!t]
\centering
\caption{Topological noise on TOPKIDS\cite{lahner2016shrec}. The table presents the mean geodesic errors ($\times 100$). \textbf{Bold}: Best. \underline{Underline}: Runner-up.}
\scalebox{1.0}{
\begin{tabular}{lrc}
\hline
                    & \multicolumn{1}{c}{TOPKIDS} & \multicolumn{1}{c}{Fully intrinsic} \\ \hline
                    \multicolumn{3}{c}{Axiomatic Methods}    \\
ZoomOut             & 33.7        & \ding{51}              \\
DiscreteOp          & 35.5        & \ding{51}               \\
SmoothShells        & 11.8        & \ding{55}              \\
\hline
                    \multicolumn{3}{c}{Unsupervised Methods}     \\
UnsupFMNet          & 38.5        & \ding{51}  \\
SURFMNet            & 48.6        & \ding{51}               \\
Deep Shells         & 13.7        & \ding{55}              \\
NeuroMorph          & 13.8        & \ding{55}            \\
AttentiveFMaps      & 23.4        & \ding{51}              \\
ULRSSM        & \underline{9.4}   & \ding{51}         \\
ULRSSM(+TTA)        & 9.2         & \ding{51}     \\
Ours                & \textbf{7.6} & \ding{51}  \\
\hline
\end{tabular}}
\label{tab: topo}
\end{table}

\section{B. Matching with topological noise} 
Topological noise presents a significant challenge for shape-matching methods as it distorts the intrinsic shape geometry in a non-isometric manner. To evaluate the robustness of our method against topological noise, we conducted experiments using the SHREC'16 TOPKIDS dataset~\cite{lahner2016shrec}, which consists of 26 shapes, one of which is noise-free, and the remaining 25 shapes contain varying degrees of topological noise.

The matching results are summarized in Table \ref{tab: topo}. Since topological noise distorts the intrinsic geometry of shapes, many purely intrinsic methods—such as SURFMNet and AttentionFMaps—struggle to produce satisfactory results under such conditions. In contrast, methods that incorporate extrinsic information, such as SmoothShell and DeepShell, often demonstrate improved robustness and accuracy. Notably, the purely intrinsic method ULRSSM still performs strongly in the presence of topological noise by simultaneously optimizing pointwise and functional maps. However, our approach introduces a novel perspective through a contrastive learning framework, which enhances both feature accuracy and robustness. As a result, it outperforms both ULRSSM and its fine-tuned variant, highlighting the superior performance and effectiveness of our method.

\begin{figure}[h!t]
	\centering
	\includegraphics[width=1\linewidth]{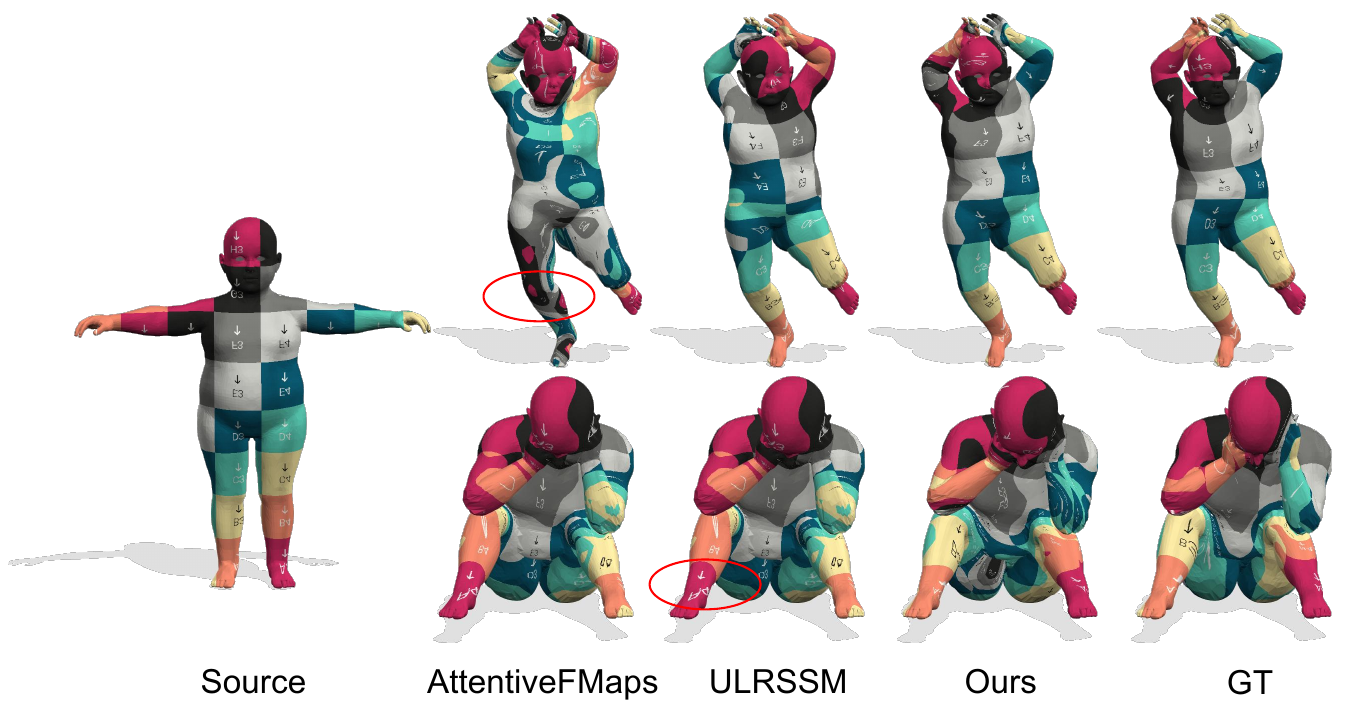}
	\caption{Comparisons with other methods on shape matching with topological noise~\cite{lahner2016shrec}. Our results, with smoother and more accurate texture distributions, illustrate that our approach is more robust to topological noise compared to existing methods
    }\label{fig7: topo}
\end{figure}

\section{C. Runtime comparison}\label{sec: runtime}
\begin{figure}[h!t]
	\centering
	\subfigure
	{
		\begin{minipage}[b]{.45\linewidth}
			\centering
			\includegraphics[scale=0.29]{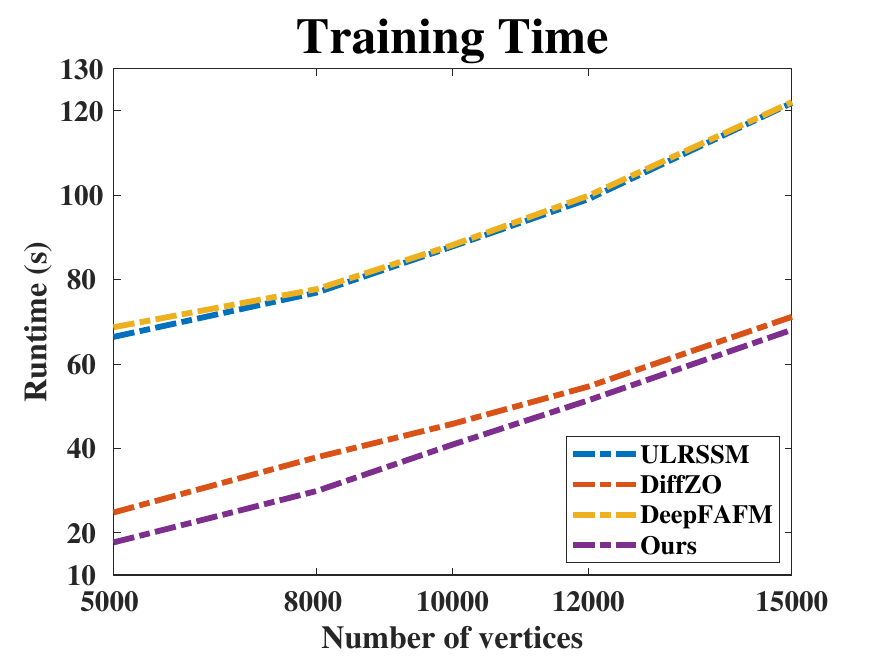}
		\end{minipage}
	}
	\subfigure
	{
		\begin{minipage}[b]{.45\linewidth}
			\centering
			\includegraphics[scale=0.29]{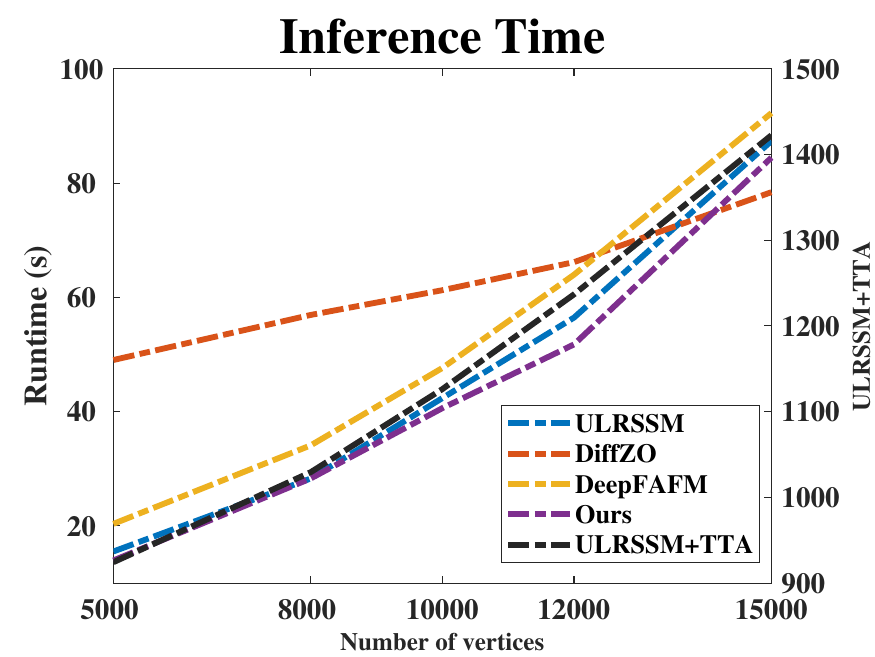}
		\end{minipage}
	}
	\caption{{Runtime comparison with different numbers of vertices. Left: Average training time per method (100 iterations). Right: Average inference time per method (100 shape pairs), where ULRSSM(+TTA) runtime (black line) is plotted against the right-hand axis due to its exceptional magnitude ($\geq$ 50× longer than other methods), the remaining competing approaches are scaled against the left-hand axis. Our method achieves best performance} }
	\label{fig: runtime}
\end{figure}

We evaluate the training and inference time of our method across datasets with varying vertex resolutions~\cite{wang2019robust}, namely 5K, 8K, 10K, 12K, and 15K vertices, and compare it to the state-of-the-art unsupervised methods. All competitors are implemented using their original parameter configurations as reported in respective publications. Quantitative comparisons are presented in Figure~\ref{fig: runtime}. 
Our method demonstrates the shortest training time, exhibiting over twice the speed of competing approaches like ULRSSM and DeepFAFM. While DiffZO currently stands as the most efficient functional maps learning method by eliminating time-consuming functional maps solvers (alongside our approach), it incurs additional computational overhead through implicit iterative solving of both pointwise and functional maps. In contrast, our framework achieves exceptional computational efficiency through: (1) the simplified functional map architecture to date, and (2) a contrastive learning framework requiring only a minimal selection of positive pairs. These innovations 
contribute substantially to the superior training efficiency of our approach.
% In inference time comparisons, our method consistently demonstrates superior performance. DiffZO experiences significantly prolonged inference latency due to its ZoomOut-style iterative upsampling for pointwise map computation. While its implicit representation benefits from the KeOps acceleration library~\cite{charlier2021kernel}, yielding superior efficiency on high-resolution meshes (e.g., 15K vertices), this does not compensate for its overall latency.
In terms of inference efficiency, our method consistently outperforms competitors. DiffZO suffers from notable latency caused by its iterative ZoomOut-style upsampling for pointwise map computation. Although its implicit representation leverages the KeOps library~\cite{charlier2021kernel} to improve handling of high-resolution meshes (e.g., 15K vertices), this acceleration fails to offset the overall computational overhead. DeepFAFM further exceeds ULRSSM's inference time due to its additional filter function computations. While sharing ULRSSM's inference strategy, our approach benefits from lower feature dimensions at both the input and output stages ( ours: input/output = 16/128; ULRSSM:  = 128/256 ), resulting in faster execution. Notably, ULRSSM's test-time adaptation (TTA) refinement strategy, while improving results, increases inference time by orders of magnitude. Crucially, without relying on any time-consuming fine-tuning techniques, our method surpasses even ULRSSM (+TTA) in final accuracy.
All the statistics were collected on a server with Intel(R) Xeon(R) Platinum 8358 CPU @ 2.60GHz, and a single NVIDIA A100-958 SXM4-80GB GPU.

\section{D. Parameter analysis}\label{sec: para}
Since the cross-similarity threshold $p_c$ and self-similarity threshold $p_s$ play a critical role in our architecture, selecting the appropriate number is essential for improving matching results. Parameter analysis experiments are conducted on the FAUST, SCAPE, SHREC'19, and SMAL datasets to explore the effects of different parameter settings on the matching accuracy and generalization performance, and the results are summarized in Table~\ref{tab: parameter analysis}. 
As the value of $p_c$ and $p_s$ increases, the results on FAUST, SCAPE, and SMAL remain largely unchanged, clearly demonstrating that our method is robust to both near-isometric and non-isometric matching. Regarding the impact of the similarity threshold on the network’s generalization ability, we observe that increasing $p_c$ and $p_s$ has a minimal effect when the network is trained on FAUST. In contrast, for SCAPE, increasing $p_c$ and $p_s$ helps mitigate the impact of inconsistent data distribution and enhances generalization performance. Finally, we set $p_c = p_s = 30 $ for near-isometric matching, anisotropic
meshing, and generalization, and $p_c = p_s = 10$ for non-isometric and topological noise matching.

\begin{table}[h!t]
\centering
\caption{Parameter analysis on remeshed FAUST, SCAPE, SHREC'19, and SMAL, respectively. }
\scalebox{1.0}{
\begin{tabular}{lrrrrc}
\hline
Train               & \multicolumn{2}{c}{F} & \multicolumn{2}{c}{S} & \multicolumn{1}{c}{\multirow{2}{*}{SMAL}}\\ \cline{2-5}
Test                & F       & S19       & S   & S19      \\ \hline
$p_c = p_s = 10$                  & 1.5     & 6.0   & 1.9       & 9.7  & 3.6  \\ 
$p_c = p_s = 20$                  & 1.5    & 6.2    & 1.9     & 8.2  & 3.7 \\ 
$p_c = p_s = 30$                  & 1.6    & 5.8   & 2.1      & 5.8  & 3.5 \\ 
$p_c = p_s = 40$                  & 1.6    & 5.0   & 2.1      & 6.3   & 3.8 \\ 
$p_c = p_s = 50$                  & 1.6    & 5.4  & 2.2      & 6.6  & 3.6 \\ 
\hline
\end{tabular}}
\label{tab: parameter analysis}
\end{table}

\section{E. Ablation Studies}\label{sec: abl}
The ablation studies we conduct on the challenging topological noise dataset (e.g., TOPKIDS) to evaluate the importance of our two losses, namely, cross- and self-contrastive losses.

The results are summarized in Table \ref{tab: ablation study}. The first and second rows show the network trained without $L_{cross}$ and $L_{self}$, respectively. Comparing the first and last rows, we observe that $L_{corss}$ significantly improves matching performance by encouraging consistency in feature representations. Similarly, comparing the second and last rows, we see that $L_{self}$ enhances matching performance by promoting the self-discriminative ability of vertex representations, leading to more informative feature embeddings.

\begin{table}[h!t]
\centering
\caption{Ablation study on TOPKIDS\cite{lahner2016shrec} dataset. }
\scalebox{1.0}{
\begin{tabular}{lcc}
\hline
               & \multicolumn{2}{c}{TOPKIDS} \\ \hline
w.o $L_{cross}$ & \multicolumn{2}{r}{17.7}     \\ %\hline
w.o $L_{self}$         & \multicolumn{2}{r}{8.4}     \\ %\hline
Ours        & \multicolumn{2}{r}{\textbf{7.6}}     \\ \hline
\end{tabular}}
\label{tab: ablation study}
\end{table}

\section{F. Qualitative Results}\label{sec: vis}
In this section, we provide the qualitative results of our method on SHREC'19, SMAL, and DT4D-H corresponding to the quantitative results reported in the main text. 

\begin{figure*}[h!t]
	\centering
	\includegraphics[width=0.9\linewidth]{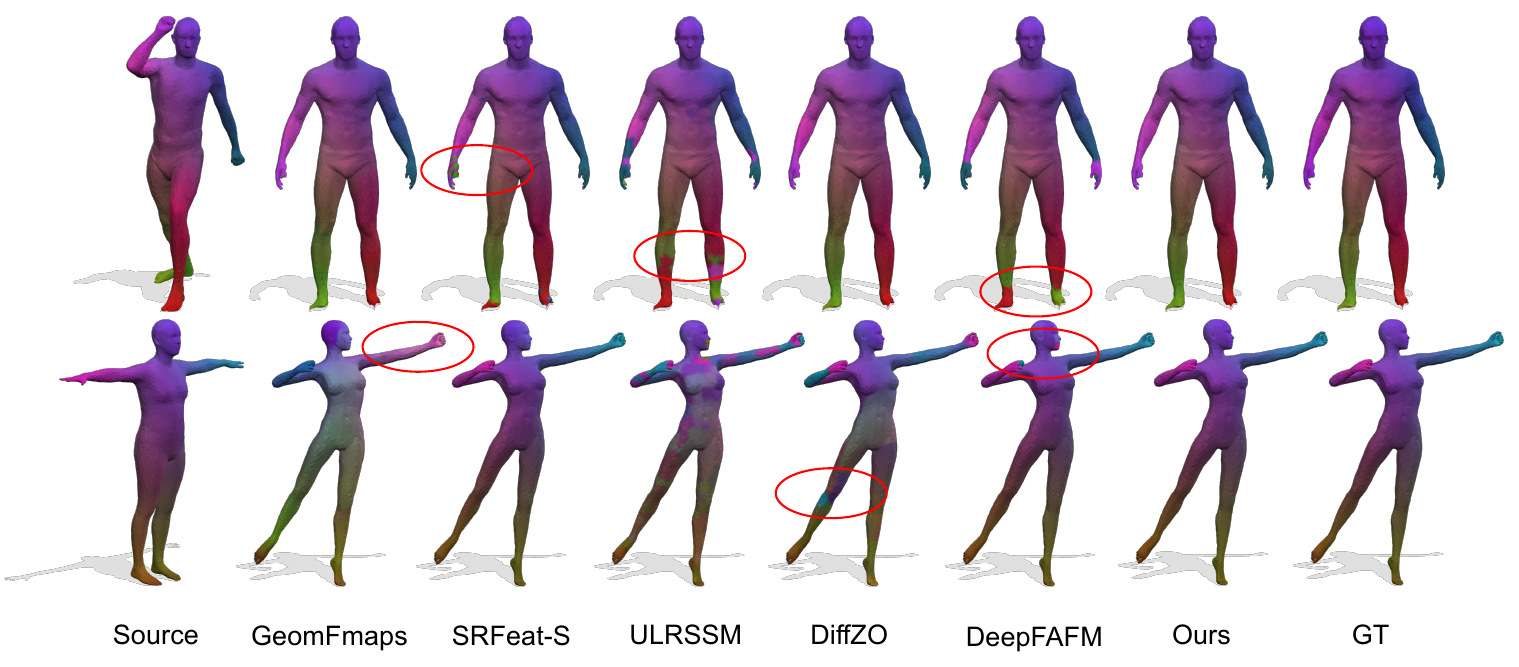}
	\caption{ Comparisons of cross-dataset generalization performance with other methods. Top row: training on FAUST~\cite{ren2018continuous} and testing on SHREC'19~\cite{melzi2019shrec}. Bottom row: Training on SCAPE~\cite{ren2018continuous} and testing on SHREC'19. Our method exhibits fewer errors and less color distortion compared to other approaches, highlighting its robust performance.    
    }
	\label{fig5: iso_gen}
\end{figure*}

\begin{figure*}[h!t]
	\centering
	\includegraphics[width=0.9\linewidth]{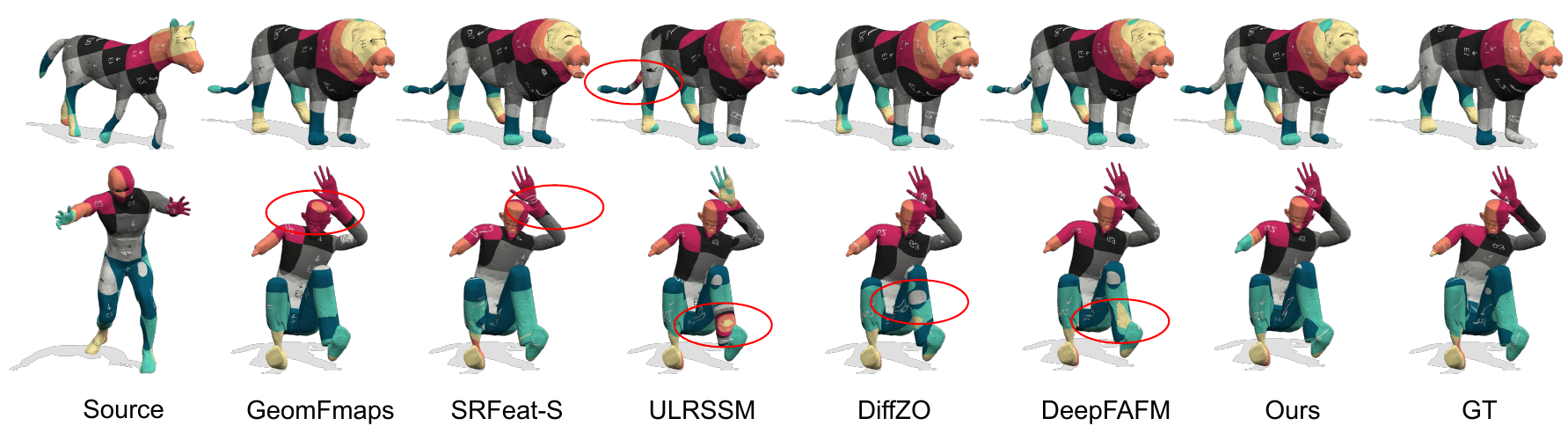}
	\caption{ Comparisons with other methods on the SMAL~\cite{zuffi20173d} (top row) and DT4D-H~\cite{2022SmoothNonRigidShapeMatchingviaEffectiveDirichletEnergyOptimization} (bottom row) datasets. Our approach results in fewer errors and less texture distortion than other methods, demonstrating its superior performance for non-isometric shape matching.}
	\label{fig6: non_iso}
\end{figure*}

\end{document}